\documentclass[twocolumn, switch]{article} 

\usepackage{preprint}
\usepackage{amsmath, amsthm, amssymb, amsfonts}

\usepackage[numbers,square]{natbib}
\usepackage[utf8]{inputenc}	
\usepackage[T1]{fontenc}	
\usepackage{xcolor}		
\usepackage[colorlinks = true,
            linkcolor = purple,
            urlcolor  = blue,
            citecolor = cyan,
            anchorcolor = black]{hyperref}	
\usepackage{booktabs} 		
\usepackage{nicefrac}		
\usepackage{microtype}		
\usepackage{lineno}		
\usepackage{float}			

\usepackage{lipsum}		
\usepackage{array}
\usepackage{textcomp}
\usepackage{multirow}
\usepackage{algorithm}
\usepackage{algorithmic}

\usepackage{pifont}
\newcommand{\cmark}{\checkmark}
\newcommand{\xmark}{\ensuremath{\times}}
\usepackage[caption=false,font=small,labelfont=bf]{subfig}
\usepackage{caption}
\usepackage{tabularx}
\usepackage[table]{xcolor}

\usepackage{newfloat}
\DeclareFloatingEnvironment[name={Supplementary Figure}]{suppfigure}
\usepackage{sidecap}
\sidecaptionvpos{figure}{c}

\usepackage{titlesec}
\titlespacing\section{0pt}{12pt plus 3pt minus 3pt}{1pt plus 1pt minus 1pt}
\titlespacing\subsection{0pt}{10pt plus 3pt minus 3pt}{1pt plus 1pt minus 1pt}
\titlespacing\subsubsection{0pt}{8pt plus 3pt minus 3pt}{1pt plus 1pt minus 1pt}

\usepackage{tikz,xcolor,hyperref}
\usepackage{orcidlink}

\definecolor{lime}{HTML}{A6CE39}
\definecolor{headergray}{gray}{0.9}
\definecolor{aqigreen}{RGB}{150,235,150}
\definecolor{aqiyellow}{RGB}{255,255,170}
\definecolor{aqiorange}{RGB}{255,200,140}
\definecolor{aqired}{RGB}{255,170,170}
\definecolor{aqipurple}{RGB}{210,180,230}
\definecolor{aqimaroon}{RGB}{230,170,180}

\DeclareRobustCommand{\orcidicon}{
	\begin{tikzpicture}
	\draw[lime, fill=lime] (0,0) 
	circle [radius=0.16] 
	node[white] {{\fontfamily{qag}\selectfont \tiny ID}};
	\draw[white, fill=white] (-0.0625,0.095) 
	circle [radius=0.007];
	\end{tikzpicture}
	\hspace{-2mm}
}
\foreach \x in {A, ..., Z}{\expandafter\xdef\csname orcid\x\endcsname{\noexpand\href{https://orcid.org/\csname orcidauthor\x\endcsname}
			{\noexpand\orcidicon}}
}

\title{M$^2$FedAQI: Multimodal Federated Learning for Air Quality Prediction on Heterogeneous Edge Devices}

\usepackage{xwatermark}
\newwatermark[firstpage,color=gray!90,angle=0,scale=0.28, xpos=0in,ypos=-5in]{*correspondence: \texttt{tamoghnaojha@iitism.ac.in}}

\usepackage{authblk}

\author[1\thanks{\tt{manjil\_nepal@srmap.edu.in}}]{Manjil Nepal}
\author[2\thanks{\tt{24im0004@iitism.ac.in}}]{Kimsie Phan}
\author[2\thanks{\tt{tamoghnaojha@iitism.ac.in}}]{Tamoghna Ojha}
\author[3\thanks{\tt{aritra.dutta@ucf.edu}}]{Aritra Dutta}
\author[1\thanks{\tt{krishnasivaprasad.m@srmap.edu.in}}]{M Krishna Siva Prasad}

\affil[1]{Department of Computer Science and Engineering, SRM University-AP, India}
\affil[2]{Department of Mathematics and Computing, IIT (ISM) Dhanbad, India}
\affil[3]{Department of Mathematics and the Department of Computer Science, University of Central Florida, USA}

\begin{document}

\twocolumn[ 
  \begin{@twocolumnfalse} 
  
\maketitle

\begin{abstract}
Accurate air quality prediction is essential for public health, environmental monitoring, and industrial safety. However, most existing approaches rely on centralized learning paradigms, which introduce challenges related to scalability, privacy preservation, and communication overhead in distributed Internet of Things (IoT) environments. Moreover, current federated learning (FL) based solutions predominantly utilize unimodal data, limiting their capability to capture complex environmental patterns. To address these limitations, we propose M$^2$FedAQI, a lightweight multimodal federated framework for decentralized Air Quality Index (AQI) prediction across heterogeneous edge devices. The proposed framework integrates visual and tabular modalities through a feature modulation based fusion mechanism that enables efficient cross-modal interaction while maintaining low computational overhead. M$^2$FedAQI is evaluated on two benchmark datasets, PM25Vision and TRAQID, for both classification and regression tasks under centralized and federated settings. Experimental results demonstrate that M$^2$FedAQI consistently outperforms existing approaches, achieving improvements of up to 11.0\% in Accuracy, 3.53\% in AUC, 12.2\% in F1-score, and 18.0\% in $R^2$, while reducing MAE and RMSE by up to 25.4\% and 20.4\%, respectively, compared with the strongest baselines. Furthermore, deployment on heterogeneous edge devices demonstrates efficient resource utilization in terms of communication overhead, memory footprint, and computational cost. To enhance communication security, TLS-based authentication is incorporated to ensure secure client participation and protect the FL communication channel from unauthorized third-party access without modifying the underlying FL protocol.
\end{abstract}

\textbf{Keywords:} Air Quality Prediction, Federated Learning, Multimodal Learning, Edge Computing, Internet of Things (IoT)

\vspace{0.35cm}

  \end{@twocolumnfalse} 
] 



\section{Introduction}
\label{sec:introduction}
Air pollution has become a major problem for both the environment and public health. There have been increasing amounts of hazardous chemicals and particulates being introduced into the atmosphere by burning fossil fuels and other human activities. These chemicals particularly include Nitrogen Dioxide (NO$_2$), Carbon Monoxide (CO), Carbon Dioxide (CO$_2$), Ozone (O$_3$), Sulfur Dioxide (SO$_2$), and fine particulate matter that can cause severe damage to the ecosystem and particularly human health, such as asthma, chronic obstructive pulmonary disease, heart disease, and lung cancer \cite{patel2023review}. The WHO estimates that about 99\% of all people live in areas with air pollution levels higher than those considered safe and containing harmful pollutants, with the highest exposure observed in low- and middle-income developing countries \cite{rentschler2023global}.

The Air Quality Index (AQI) is a key indicator used to determine the level of air pollution in the atmosphere, and as this level increases, so does the public health risk. AQI measures overall air quality through a standardized scale that runs from 0 to 500, which is further divided into categories representing different pollution levels, ranging from excellent and good to light–moderate, moderate, bad, and heavily polluted conditions \cite{horn2024air}. With the rapid growth of Internet of Things (IoT) based AQI monitoring systems, an enormous volume of multimodal data has been generated in various forms, including images, text, audio, and video \cite{zhang2023recent}. As a result, researchers over the years have put forward numerous solutions aimed at forecasting and controlling air pollution through IoT and AI techniques \cite{garcia2025advancements, gangwar2023state}. At the same time, the growing availability of multimodal AQI datasets has opened up considerable opportunities for building advanced artificial intelligence (AI) models that can capture the complex relationships between environmental factors and atmospheric pollution \cite{han2025pm25vision, kathalkar2024traqid}.

The existing AQI systems are based on the centralized architecture in which the sensor data of several monitoring stations are sent to the central servers, where it is processed and analyzed \cite{jin2021multivariate}. There are several limitations with a centralized approach: (i) transferring high-frequency multi-parameter data from distributed sensors to a central server increases bandwidth usage and communication overhead, (ii) data storage requires monitoring information from multiple stations to be accumulated in one location, raising privacy and security concerns, (iii) centralized systems face scalability bottlenecks as increasing numbers of monitoring stations generate larger data volumes, and (iv) centralized processing requires substantial computational and storage resources at the cloud server to manage continuously growing AQI data streams. \cite{yarham2024enhancing, vahabi2025federated}. While some address this by the use of FL as a training technique, many of these approaches have been evaluated primarily in controlled or simulated environments, with relatively few works demonstrating end‑to‑end deployment on real AQI‑IoT infrastructure \cite{shojafar2021guest, xu2024defedtl}.

To address these challenges, we propose a lightweight multimodal architecture and employ it within a federated learning setup for privacy-preserving air quality prediction on heterogeneous edge devices. 

The main contributions of this work are as follows:

\begin{itemize}

\item We propose M$^2$FedAQI, a lightweight multimodal model for decentralized AQI prediction that integrates visual images and structured sensor readings through a feature modulation--based fusion mechanism, enabling efficient cross-modal interaction with low computational overhead.

\item We comprehensively evaluate M$^2$FedAQI on two benchmark datasets, PM25Vision and TRAQID, for both regression and classification tasks under centralized and federated settings, where the proposed model consistently outperforms unimodal and existing multimodal baselines across multiple evaluation metrics.

\item We demonstrate the robustness of M$^2$FedAQI in realistic federated environments with heterogeneous edge devices and non-IID data distributions, while also validating its practical deployment feasibility through detailed edge profiling on heterogeneous edge devices.

\item We further enhance system security via TLS-based client authentication to ensure secure client participation and protected communication without modifying the underlying federated learning protocol.

\end{itemize}

The rest of this paper is structured in the following way: Section \ref{sec:literature} includes a literature survey covering AQI prediction approaches in both centralized and federated setting using unimodal and multimodal dataset. Section \ref{sec:methodology} outlines the problem formulation, federated learning, and proposed multimodal architecture. Section \ref{sec:experimentation} contains the dataset description, evaluation metrics.
Section \ref{sec:results_analysis} contains the experimental results and discussion with the description of the datasets, performance evaluation measure. Section \ref{sec:conclusion} concludes the paper by summarizing the key findings and outlining potential directions for future work.

\section{Literature Review}
\label{sec:literature}
\subsection{Centralized Unimodal AQI Prediction Approaches}
Several studies have explored centralized unimodal AQI prediction, where data from a single modality or sensor source is collected and processed through centralized machine learning (ML) and deep learning (DL) frameworks \cite{nilesh2022iot, zhang2020deep, han2025pm25vision, mondal2024uncovering}. Kataria \textit{et al.}  \cite{kataria2022ai} proposed an IoT-enabled AQI prediction framework using sensor nodes integrated with CO and PM$_{2.5}$ and cloud-based data collection. The work proposed a Convolutional Neural Network–Long Short‑Term Memory (CNN–LSTM) with Bayesian optimization algorithm model, which achieved the best predictive performance in terms of AQI accuracy. Sarkar \textit{et al.} \cite{sarkar2022air} investigated AQI prediction using a combination of LSTM and Gated Recurrent Unit (GRU) as a proposed and compared with LSTM, GRU, Linear Regression, K-Nearest Neighbor (KNN), and Support Vector Machine (SVM). The hybrid LSTM-GRU model achieved improved prediction performance in MAE and $R^2$ metrics.

\subsection{Centralized Multimodal AQI Prediction Approaches}
Recently, multimodal learning has been investigated to enhance AQI prediction through the incorporation of heterogeneous environmental data in centralized architectures \cite{kalajdjieski2020air}. Hameed \textit{et al.}~\cite{hameed2023deep} integrated visual information extracted from CCTV images with environmental sensor data. The study proposed variants of the LSTM model that include Bi-directional LSTM, CNN-LSTM, and Convolutions LSTM to capture spatial and temporal pollution patterns from multimodal data. Among the evaluated models, these variants improved short-term, medium-term, and long-term periods over the ARIMA model, respectively. Similarly, Lilhore \textit{et al.}~\cite{lilhore2025advanced} developed a hybrid model that fused satellite imagery with temporal dynamics in pollutant and weather data by combining CNN, Bidirectional LSTM (BiLSTM), attention mechanisms, GNNs, and Neural ODEs to produce high accuracy predictions in centralized settings. The proposed model outperformed the existing model in the regression task. 

\subsection{Federated Unimodal AQI Prediction Approaches}
FL has been increasingly adopted by distributed air quality monitoring, where it is used to circumvent privacy limitations of centralized data collection. Models are trained collaboratively while data stays on each local device. Recently, Dey \textit{et al.}~\cite{dey2026greenedge} proposed GreenEdge AI, a sustainable FL framework for AQI prediction that integrates a green-aware custom LSTM (GA-CLSTM) model with energy-aware training, adaptive aggregation, and a hybrid loss function. Their approach incorporates explicit energy optimization metrics and demonstrates significant improvements in prediction accuracy while reducing energy consumption and communication overhead. In addition, Hu \textit{et al.} \cite{hu2024feddeep} proposed FedDeep, a federated deep learning framework for edge-assisted multi-urban PM2.5 forecasting that enables collaborative model training across geographically distributed monitoring sites that enables effective knowledge sharing across distributed urban regions while improving prediction accuracy and efficiency.

\subsection{Federated Multimodal AQI Prediction Approaches}
Combining multiple data modalities within a federated setup remains one of the least explored directions in the AQI literature. Rahman \textit{et al.} \cite{rahman2025multimodal}, this gap is addressed by developing a multimodal FL framework with personalization that uses drone-captured aerial imagery and ground sensor data across distributed clients. The method enables collaborative training by sharing encoder models among clients while constructing a high-quality decoder at a ground station, allowing for effective integration of multimodal data while maintaining global consistency, achieving performance improvements of approximately 2.46\% over prior baseline methods.  Arjun \textit{et al.} \cite{arjun2025multimodal} developed a multimodal FL framework to preserve privacy for the prediction of carbon footprints in smart cities, integrating various urban data sources such as air quality, energy use and transportation metrics. By incorporating differential privacy, their method ensures secure collaborative learning across cities while achieving competitive predictive performance, with a reported MAE of $3.247$ tons CO$_{2}$ equivalent, and enabling large-scale sustainability insights.

Despite significant progress in AQI prediction, several critical gaps remain unaddressed. First, most existing approaches rely on centralized architectures, leading to privacy risks, high communication overhead, and poor scalability in distributed IoT environments. Second, while FL has been explored, prior works predominantly focus on unimodal data, limiting their ability to capture complementary information from heterogeneous sources. Third, existing multimodal methods are largely centralized, failing to leverage the benefits of decentralized training while preserving data locality. Fourth, current federated multimodal approaches lack efficient cross-modal interaction mechanisms, often relying on simplistic fusion strategies that do not fully exploit modality dependencies. Finally, there is a lack of validation on resource-constrained edge devices under realistic non-IID settings, leaving practical feasibility largely unexplored, as most of the existing works are based on simulation. Therefore, this work tries to overcome these limitations by designing M$^2$FedAQI, a lightweight multimodal model that can be trained efficiently, can be practically feasible, and can be scaled for deployment in heterogeneous edge environments. An overall of the existing work has been summarized in the Table~\ref{tab:lit_comparison}.

\begin{table*}[t]
\centering
{\fontsize{8}{10}\selectfont
\caption{Comparison of Existing AQI Prediction Approaches}
\label{tab:lit_comparison}
\renewcommand{\arraystretch}{1.4}
\setlength{\tabcolsep}{5pt}
\begin{tabular}{|l|c|c|c|c|c|c|c|}
\hline
\centering
\textbf{Paper} & \textbf{Federated} & \textbf{Centralized} & \textbf{Multimodal} & \textbf{Edge Training} & \textbf{Non-IID} & \textbf{Secure Communication} & \textbf{Light Weight} \\
\hline
Kataria \textit{et al.} \cite{kataria2022ai} & \xmark & \cmark & \xmark & \cmark  & \xmark & \xmark & \xmark  \\
\hline
Sarkar \textit{et al.} \cite{sarkar2022air} & \xmark & \cmark & \xmark & \xmark  & \xmark & \xmark & \xmark  \\
\hline
Nilesh \textit{et al.} \cite{nilesh2022iot} & \xmark & \cmark & \cmark & \cmark  & \xmark & \xmark & \xmark  \\
\hline
Zhang \textit{et al.} \cite{zhang2020deep} & \xmark & \cmark & \xmark & \xmark  & \xmark & \xmark & \xmark  \\
\hline
Hameed \textit{et al.} \cite{hameed2023deep} & \xmark & \cmark & \cmark & \xmark  & \xmark & \xmark & \xmark  \\
\hline
Lilhore \textit{et al.} \cite{lilhore2025advanced} & \xmark & \cmark & \cmark & \xmark  & \xmark & \xmark & \xmark  \\
\hline
Yang Han \textit{et al.} \cite{han2025pm25vision} & \xmark & \cmark & \xmark & \xmark  & \xmark & \xmark & \xmark  \\
\hline
Mondal \textit{et al.} \cite{mondal2024uncovering} & \xmark & \cmark & \xmark & \xmark  & \xmark & \xmark & \cmark  \\
\hline
Kalajdjieski \textit{et al.} \cite{kalajdjieski2020air} & \xmark & \cmark & \cmark & \xmark  & \xmark & \xmark & \xmark  \\
\hline
Dey \textit{et al.} \cite{dey2026greenedge} & \cmark & \cmark & \xmark & \xmark  & \cmark & \xmark & \cmark  \\
\hline
Hu \textit{et al.} \cite{hu2024feddeep} & \cmark & \cmark & \xmark & \xmark  & \cmark & \xmark & \xmark  \\
\hline
Rahman \textit{et al.} \cite{rahman2025multimodal} & \cmark & \cmark & \cmark & \xmark  & \cmark & \xmark & \xmark  \\
\hline
Arjun \textit{et al.} \cite{arjun2025multimodal} & \cmark & \cmark & \cmark & \xmark  & \xmark & \cmark & \xmark  \\
\hline
\textbf{M$^2$FedAQI} & \cmark & \cmark & \cmark & \cmark  & \cmark & \cmark & \cmark  \\
\hline
\end{tabular}
}
\end{table*}

\section{Methodology}\label{sec:methodology}
\subsection{Problem Formulation}
The objective of air quality prediction is to predict the Air Quality Index (AQI) based on fine particulate matter (PM$_{2.5}$) concentration with the use of heterogeneous environmental data based on distributed sensing systems. In our context, an edge device at a single location gathers multimodal data in the form of visual data and structured environmental data.

Formally, let $\mathcal{D}_k = \{(x_i^{img}, x_i^{tab}, y_i)\}_{i=1}^{N_k}$ denote the local dataset stored on client $k$, where $x_i^{img}$ represents the input image capturing atmospheric conditions, $x_i^{tab}$ denotes the corresponding tabular environmental features, and $y_i$ represents the ground-truth AQI value or level. The total number of participating clients is denoted by $K$.

The goal is to learn a multimodal prediction model for two downstream task: classification and regression.
\begin{equation}
f_{w}(x^{img}, x^{tab}) \rightarrow y
\end{equation}
where $w$ represents the model parameters and $y$ denotes the predicted air quality indicator.

As mentioned previously, we consider two downstream settings:

\textbf{Regression Task:}
The goal of this task is to predict a continuous PM$_{2.5}$ value:
\begin{equation}
y \in \mathbb{R}
\end{equation}
The regression objective minimizes the mean absolute error (MAE) between predicted and ground-truth AQI values:

\begin{equation}
\mathcal{L}_{regression} = \frac{1}{N}\sum_{i=1}^{N} |y_i - \hat{y}_i|
\end{equation}

where $\hat{y}_i = f_{w}(x_i^{img}, x_i^{tab})$ is the predicted AQI.\\

\textbf{Classification Task:}
Alternatively, AQI may be defined in the form of discrete air quality levels which are defined on predetermined levels. According to Table~\ref{tab:aqi_levels}, the AQI scale has several categories that are decided according to the level of air quality. The model in this case is a predictor of a categorical label:

\begin{equation}
y \in \{1,2,\dots,C\}
\end{equation}

where $C$ denotes the number of AQI categories, which is six in our case.

The classification objective is optimized using the cross-entropy loss:

\begin{equation}
\mathcal{L}_{classification} = -\sum_{i=1}^{N} y_i \log(\hat{y}_i)
\end{equation}

Conclusively, the general goal is to learn model parameters, denoted as $W$, that are optimal for the minimization of the task-specific loss.

\begin{table}[t]
\centering
{\fontsize{7.5}{10}\selectfont
\caption{Air Quality Index (AQI) Categories}
\label{tab:aqi_levels}
\renewcommand{\arraystretch}{1.3}
\setlength{\tabcolsep}{4pt}
\footnotesize

\begin{tabular}{|c|m{1.96cm}|c|m{3.5cm}|}
\hline
\textbf{Color} & \textbf{Category} & \textbf{AQI} & \textbf{Health Implications} \\
\hline

\centering\textcolor{green!70!black}{\rule{0.4cm}{0.4cm}} 
& Good & 0--50 
& Air quality is satisfactory; negligible health risk. \\
\hline

\centering\textcolor{yellow!80!black}{\rule{0.4cm}{0.4cm}} 
& Moderate & 51--100 
& Acceptable; slight risk for sensitive individuals. \\
\hline

\centering\textcolor{orange}{\rule{0.4cm}{0.4cm}} 
& Unhealthy for Sensitive Groups 
& 101--150 
& Sensitive groups may experience health effects. \\
\hline

\centering\textcolor{red}{\rule{0.4cm}{0.4cm}} 
& Unhealthy 
& 151--200 
& General population may experience adverse effects. \\
\hline

\centering\textcolor{purple}{\rule{0.4cm}{0.4cm}} 
& Very Unhealthy 
& 201--300 
& Health alert; increased risk for all individuals. \\
\hline

\centering\textcolor{brown}{\rule{0.4cm}{0.4cm}} 
& Hazardous 
& 301+ 
& Serious health effects; emergency conditions. \\
\hline
\end{tabular}
}
\end{table}

\subsection{Federated Learning}
\begin{figure}[!h]
\centering
\includegraphics[width=0.95\columnwidth]{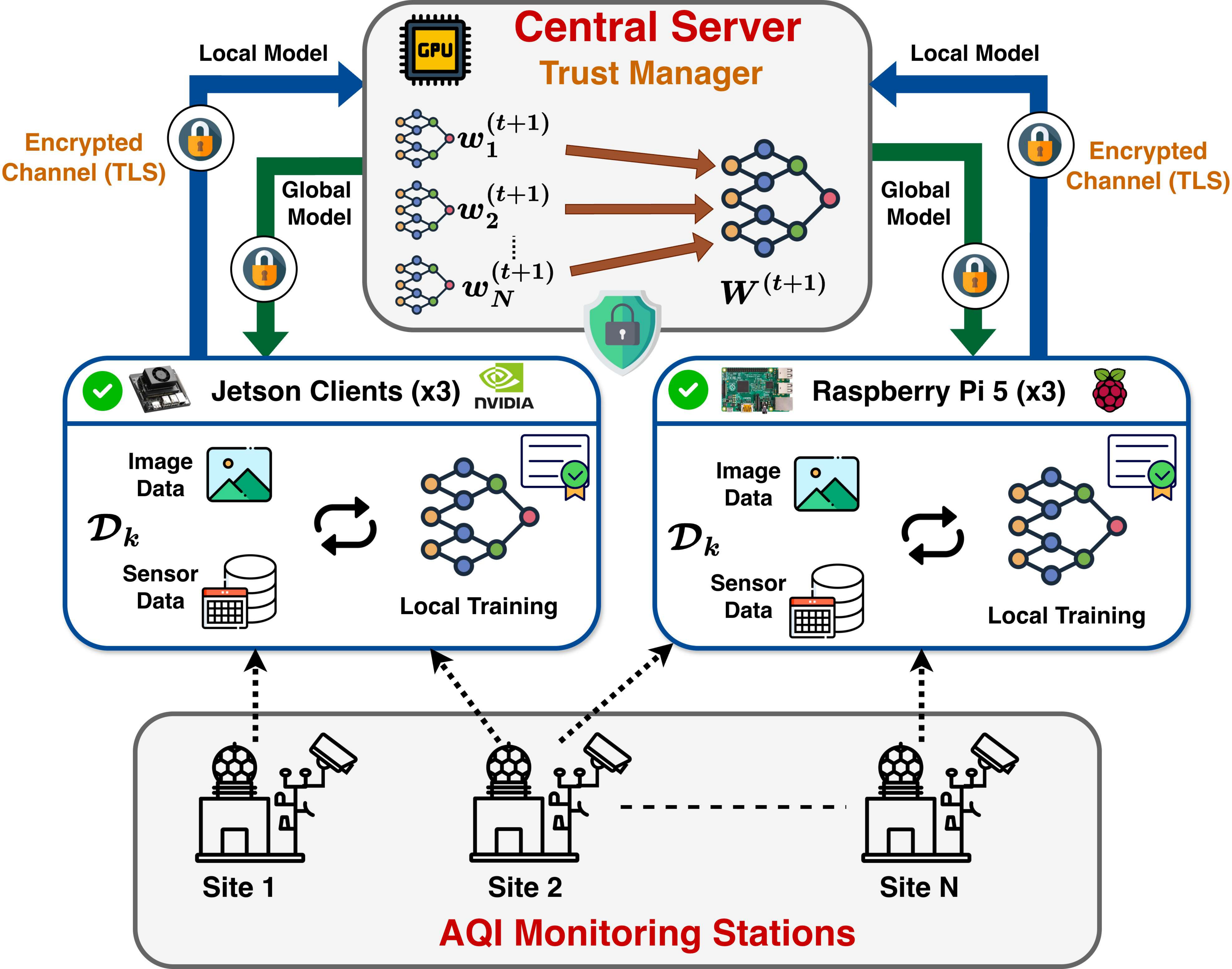}
\caption{Proposed Federated Learning Process}
\label{fig:proposed_fl}
\end{figure}

We adopt a synchronous federated learning (FL) paradigm to enable collaborative training across distributed edge clients while preserving data locality. As illustrated in Fig.~\ref{fig:proposed_fl}, a set of $k$ clients participate in training by maintaining local datasets $\mathcal{D}_k$ and optimizing a shared global model without exchanging raw data.

At communication round $t$, the server broadcasts the current global model parameters $W^{(t)}$ to all the participating clients. Each client $k$ performs local optimization over its dataset $\mathcal{D}_k$ for $E$ epochs, minimizing the local objective:
\begin{equation}
\mathcal{L}_k(W) = \frac{1}{|\mathcal{D}_k|} \sum_{(x_i, y_i) \in \mathcal{D}_k} \ell\big(f_{W}(x_i), y_i\big),
\end{equation}
where $\ell(\cdot)$ denotes the task-specific loss function.

After local training, each client obtains updated parameters $w_k^{(t+1)}$, which are transmitted to the central server. The server aggregates these updates using the standard Federated Averaging (FedAvg) algorithm \cite{mcmahan2017communication}:
\begin{equation}
W^{(t+1)} = \sum_{k=1}^{K} \frac{|\mathcal{D}_k|}{\sum_{j=1}^{K} |\mathcal{D}_j|} \, w_k^{(t+1)}.
\end{equation}

This iterative process continues for multiple communication rounds until convergence. The FL framework enables scalable and privacy-preserving learning under non-IID data distributions commonly observed across geographically distributed monitoring stations. 

To ensure secure participation, the communication between clients and the server is secured with the help of Transport Layer Security (TLS), which provides the encryption, authentication and integrity of model updates. This allows unauthorized access and tampering during transmission, allowing reliable deployment in a real-world edge environment without making changes to the underlying federated learning protocol. The overall federated learning workflow is summarized in Algorithm~\ref{alg:fedavg_tls}.

\begin{algorithm}
\caption{FedAvg with TLS}
\label{alg:fedavg_tls}
\begin{algorithmic}[1]
\REQUIRE Global model $W^{(0)}$, rounds $T$, local epochs $E$

\STATE Register and authenticate clients via secure credentials
\STATE Establish TLS-secured communication channels

\FOR{$t = 0, \dots, T-1$}
    \STATE Server selects clients $\mathcal{S}_t$ (synchronous participation)
    \STATE Broadcast $W^{(t)}$ to $\mathcal{S}_t$ over secure channels

    \FOR{each client $k \in \mathcal{S}_t$ \textbf{in parallel}}
        \STATE Update $w_k$ via local training on $\mathcal{D}_k$ for $E$ epochs
        \STATE Send updated $w_k$ to server via TLS
    \ENDFOR

    \STATE Aggregate:
    \STATE $W^{(t+1)} \leftarrow \sum_{k \in \mathcal{S}_t} \frac{|\mathcal{D}_k|}{\sum_{j \in \mathcal{S}_t} |\mathcal{D}_j|} w_k$

    \STATE Evaluate global model using client-side validation
\ENDFOR

\RETURN $W^{(T)}$
\end{algorithmic}
\end{algorithm}

\subsection{Proposed Multimodal Architecture: $M^2FedAQI$}
\begin{figure*}[t]
\centering
\includegraphics[width=1.8\columnwidth]{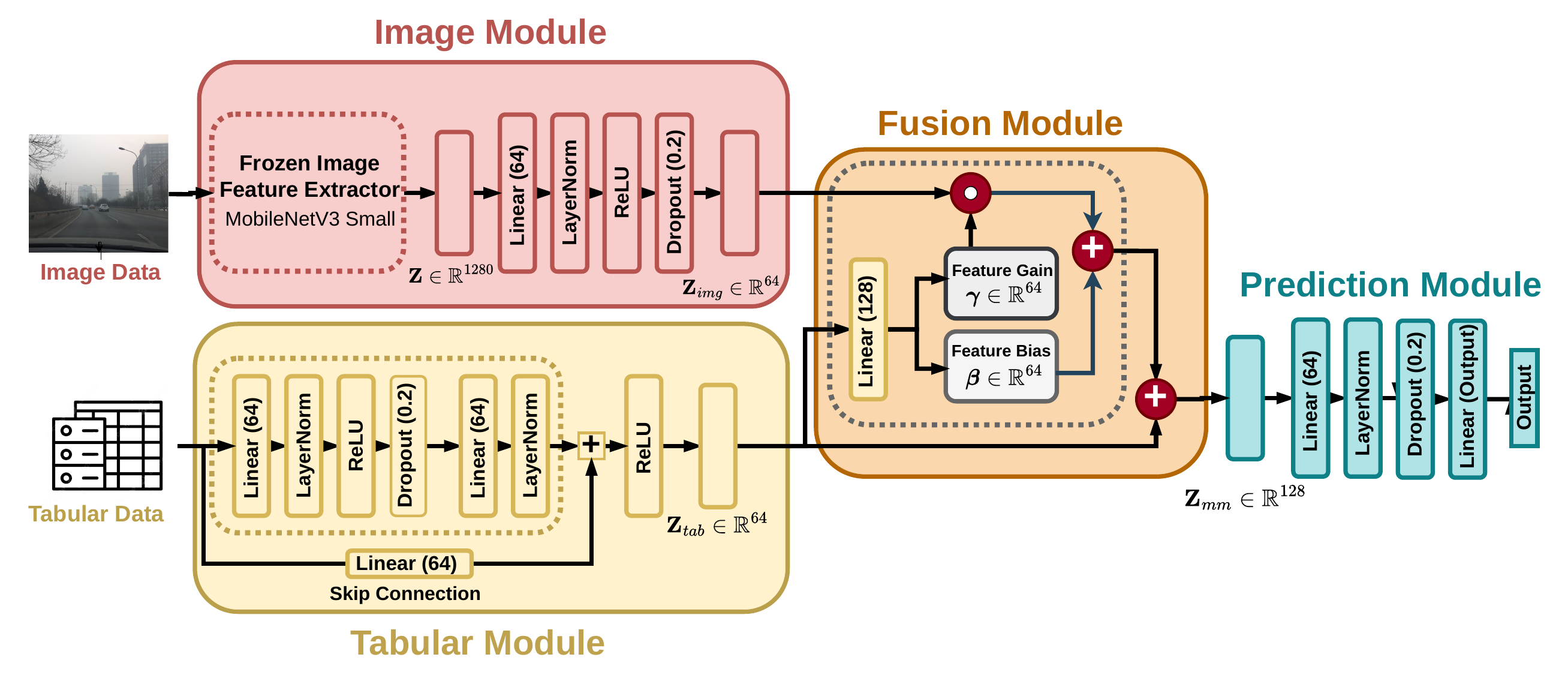}
\caption{Proposed Multimodal Architecture (M$^2$FedAQI)}
\label{fig:proposed_mm}
\end{figure*}

M$^2$FedAQI combines heterogeneous sources of data by integrating the visual information and structured environmental features using a multimodal paradigm of learning. As shown in Figure~\ref{fig:proposed_mm}, M$^2$FedAQI has four primary modules, namely: \emph{image module} for feature extraction of the image modality, \emph{tabular module} for sensor data processing, \emph{fusion module} that fuses the two modalities, and lastly, the \emph{prediction module} that performs the intended downstream task. \\

\subsubsection{Image Module}
The image module is a visual processing model that uses a frozen pretrained MobileNetV3-Small \cite{howard2019searching} as the base feature extractor. MobileNetV3-Small is a convolutional architecture whose layers are frozen to decrease the complexity of training and minimize its execution on the edge devices with resource constraints. The image representation extracted $Z \in \mathbb{R}^{1280}$ is then projected using a Linear layer, Layer Normalization, ReLU activation and Dropout regularization. This produces a low-dimensional visual feature embedding $Z_{\text{img}} \in \mathbb{R}^{64}$ that captures high-level semantic information relevant to atmospheric conditions and environmental visibility.

\subsubsection{Tabular Module}
The tabular module works with structured environmental data with a lightweight multilayer perceptron (MLP). Two Linear layers with Layer normalization, ReLU activation, and Dropout are applied to the input tabular features to generate a low-dimensional representation of the input data. Also, a residual skip connection is added so as to stabilize training and retain the low-level feature data. The resulting tabular embedding is represented as $Z_{\text{tab}} \in \mathbb{R}^{64}$, which captures contextual environmental attributes associated with air quality conditions. 

\subsubsection{Fusion Module}
We adopt a feature modulation mechanism inspired by FiLM~\cite{perez2018film} to enable interaction between visual and tabular modalities. Specifically, the tabular embedding is passed through a linear transformation to produce two modulation vectors: a feature gain $\gamma \in \mathbb{R}^{64}$ and a feature bias $\beta \in \mathbb{R}^{64}$. These vectors adaptively transform the visual feature representation through an affine operation defined as

\begin{equation}
Z_{\text{mod}} = \gamma \odot Z_{\text{img}} + \beta ,
\end{equation}

where $\odot$ denotes element-wise multiplication. This operation allows the tabular features to dynamically reweight and adjust the visual feature channels, enabling cross-modal interaction between the two modalities. We adopt this direction of modulation (tabular-to-visual) based on preliminary experiments, which showed improved performance compared to the reverse configuration (visual-to-tabular). The resulting modulated visual representation is subsequently concatenated with the tabular embedding $Z_{\text{tab}} \in \mathbb{R}^{64}$ to produce the final multimodal feature vector $Z_{\text{mm}} \in \mathbb{R}^{128}$.

\subsubsection{Prediction Module}
Lastly, the resulting fused multimodal representation of the input,  $Z_{\text{mm}} \in \mathbb{R}^{128}$ is passed through the prediction module with a Linear layer, Layer Normalization, ReLU activation, and Dropout regularization, and a final linear layer with an output prediction that is either classification or regression task.

\section{Experimentation Details}\label{sec:experimentation}
\subsection{Dataset Description}
\subsubsection{PM25Vision}
We utilize the PM25Vision dataset for our experiments, a large-scale benchmark for visual air quality estimation that aligns geotagged street-level images with corresponding PM2.5 measurements. The dataset is constructed by matching mapillary images with air quality records from the World Air Quality Index (WAQI) monitoring network. It contains over 11,000 image–PM2.5 pairs from more than 3,200 monitoring stations worldwide, collected over a decade. Both regression and classification (AQI category) tasks are supported by each sample having a street-view image and its corresponding PM2.5 value.The dataset spans diverse geographic regions and pollution levels, enabling robust evaluation of vision-based and multimodal air quality models \cite{han2025pm25vision}. 

\subsubsection{TRAQID}
TRAQID (Traffic-Related Air Quality Image Dataset) is a multimodal dataset of large scale such that it is used to estimate air quality based on images. It includes 26,678 front and rear traffic shots, which were captured in several seasons in Hyderabad and Secunderabad, India, as well as the co-located weather parameters, particulate matter (PM) levels, and AQI measurements. The dataset captures diverse environmental conditions, including day and night scenarios and unstructured traffic patterns, across six AQI categories ranging from ``Good'' to ``Severe'', providing a comprehensive benchmark for image-based AQI prediction tasks \cite{kathalkar2024traqid}.

\subsection{Evaluation Metrics}
In order to fully assess M$^2$FedAQI, we utilize standard evaluation metrics of both regression and classification problems.

\subsubsection{Regression Metrics}
For continuous PM$_{2.5}$ prediction, we use the following regression metrics:

\textbf{Mean Absolute Error (MAE):}
\begin{equation}
MAE = \frac{1}{N} \sum_{i=1}^{N} |y_i - \hat{y}_i|
\end{equation}

\textbf{Root Mean Squared Error (RMSE):}
\begin{equation}
RMSE = \sqrt{\frac{1}{N} \sum_{i=1}^{N} (y_i - \hat{y}_i)^2}
\end{equation}

\textbf{Coefficient of Determination ($R^2$):}
\begin{equation}
R^2 = 1 - \frac{\sum_{i=1}^{N} (y_i - \hat{y}_i)^2}
{\sum_{i=1}^{N} (y_i - \bar{y})^2}
\end{equation}

\noindent where $y_i$ and $\hat{y}_i$ denote the ground truth and predicted values, respectively, $\bar{y}$ is the mean of ground truth values, and $N$ is the total number of samples.

\subsubsection{Classification Metrics}
For AQI category prediction, we report:

\textbf{Accuracy:}
\begin{equation}
Accuracy = \frac{TP + TN}{TP + TN + FP + FN}
\end{equation}

\textbf{F1-score:}
\begin{equation}
F1 = \frac{2 \cdot Precision \cdot Recall}
{Precision + Recall}
\end{equation}

\textbf{AUC (Area Under the ROC Curve):}
\begin{equation}
\text{AUC} = \int_{0}^{1} TPR(FPR) \, d(FPR)
\end{equation}

\noindent where $TP$, $TN$, $FP$, and $FN$ denote true positives, true negatives, false positives, and false negatives, respectively. Precision is defined as $Precision = \frac{TP}{TP + FP}$, and Recall (True Positive Rate) is defined as $TPR = \frac{TP}{TP + FN}$. The False Positive Rate is given by $FPR = \frac{FP}{FP + TN}$. \\

These metrics are particularly important due to class imbalance across AQI categories.

\subsection{Implementation Details}
\begin{figure}[H]
\centering
\includegraphics[width=0.95\columnwidth]{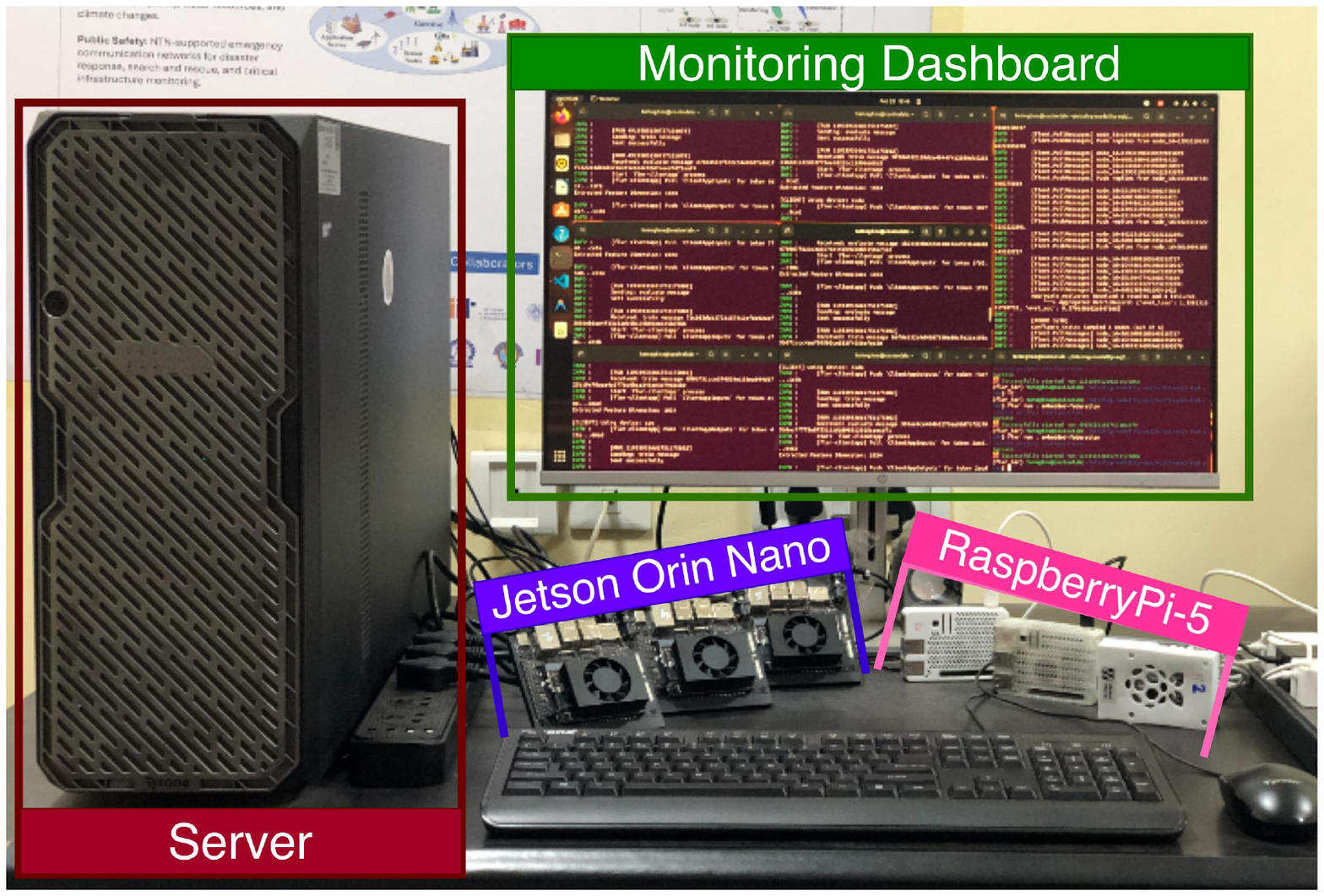}
\caption{Experimental Setup}
\label{fig:setup}
\end{figure}

M$^2$FedAQI was implemented using PyTorch \cite{paszke2019pytorch} for centralized training and the Flower framework \cite{beutel2020flower} for federated learning. For the centralized training we utilized NVIDIA RTX 4500 Ada whereas for the federated training we used six heterogeneous edge devices (three Raspberry Pi-5 and three NVIDIA Jetson Orin Nano, each with 8 GB RAM) and a central server equipped with an NVIDIA RTX 4500 Ada GPU for aggregation. The Jetson Orin Nano was configured in maximum power mode with GPU acceleration enabled, creating computational heterogeneity with the Raspberry Pi-5, which operates solely on CPU. A wireless local network (WiFi) was used with a peak bandwidth of 30Mbps to connect all the devices and enable distributed training across edge devices. The experimental setup is shown in Fig.~\ref{fig:setup}.

When training in a centralized setting, we set the learning rate to 0.01 for PM25Vision and 0.001 for TRAQID, with a batch size of 32 and a total of 50 training epochs. Training was carried out in the federated setting with 50 server rounds in which the learning rate of 0.0005 and batch size of 32 are used in both datasets in which each client completes 5 local training epochs before sending the updates to the server for aggregation. We use a synchronous federated learning model where all clients engage in each round, and the global model is updated regularly and consistently. To simulate realistic data heterogeneity across clients, non-IID data distributions are generated using a Dirichlet partitioning strategy \cite{yurochkin2019bayesian} with a concentration parameter of $\alpha = 0.5$, resulting in highly heterogeneous data distributions across the participating clients.

\begin{table}[!h]
    \centering
    \caption{Experimental Setup Configuration}
    \label{tab:experiment_setup}
    \small
    \renewcommand{\arraystretch}{1.2}
    \setlength{\tabcolsep}{8pt}
    \begin{tabular}{|>{\bfseries\raggedright\arraybackslash}p{5.3cm}|p{2.4cm}|}
        \hline
        \rowcolor{gray!15} \multicolumn{2}{|c|}{\textbf{Software Configuration}} \\
        \hline
        Deep Learning Framework & PyTorch 2.0.1 \cite{paszke2019pytorch} \\
        \hline
        Federated Learning Framework & Flower 1.29.0 \cite{beutel2020flower} \\
        \hline
        \rowcolor{gray!15} \multicolumn{2}{|c|}{\textbf{Hardware Configuration}} \\
        \hline
        Client:  3 x Raspberry Pi-5 & 8GB RAM \\
        \hline
        Client:  3 x Nvidia Jetson Orin Nano & 8GB RAM \\
        \hline
        Server: NVIDIA RTX 4500 Ada Gen & 24GB VRAM\\
        \hline
    \end{tabular}
\end{table}

\subsection{TLS Certification}
\begin{figure}[H]
\centering
\includegraphics[width=0.9\columnwidth]{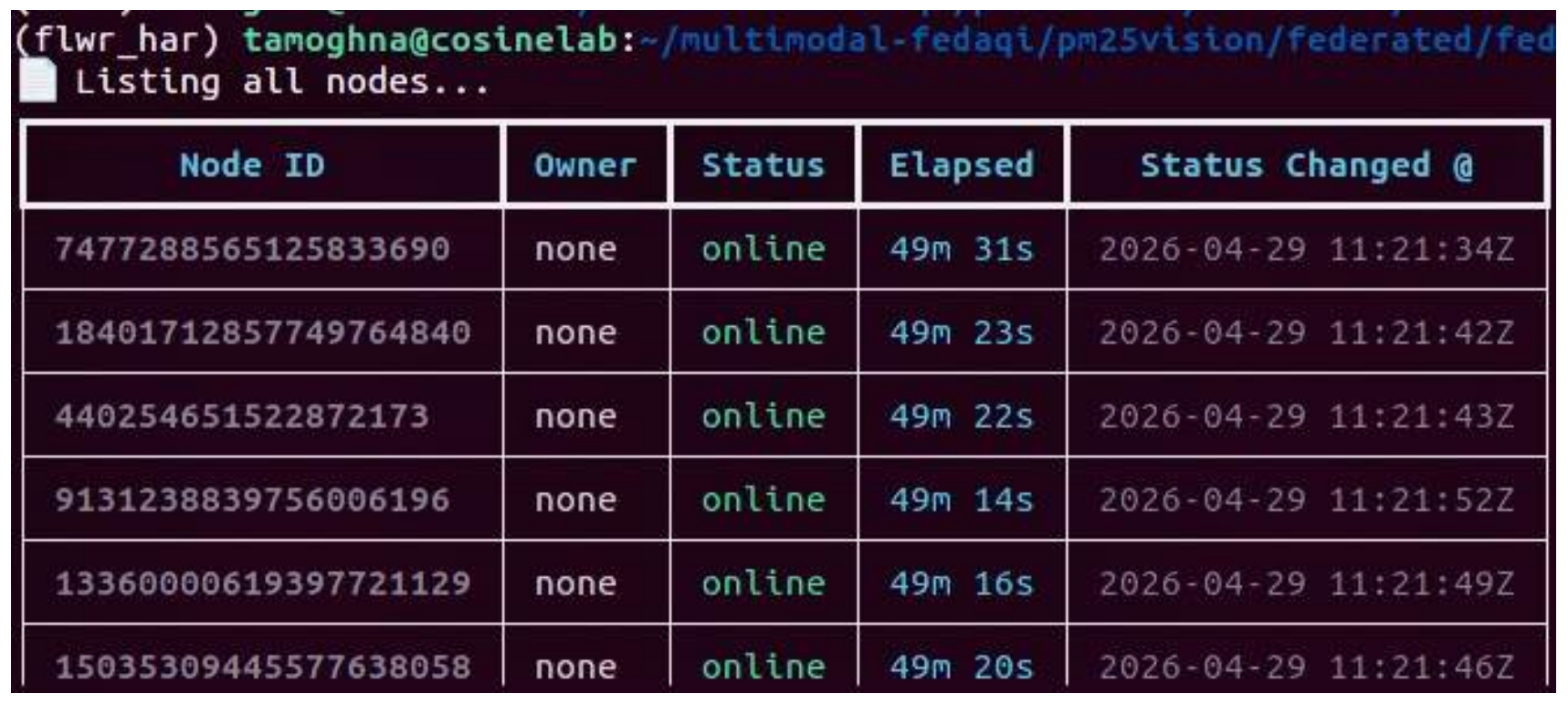}
\caption{TLS Certifcation}
\label{fig:tls}
\end{figure}

To ensure secure participation in the federated learning environment, TLS-based client authentication was employed to restrict federation access exclusively to authorized edge devices. Each client device was authenticated prior to joining the federation using securely registered credentials, as illustrated in Fig.~\ref{fig:tls}. This authentication mechanism provides encrypted and trusted communication between clients and the central server, thereby preventing unauthorized access and potential third-party attacks on the communication channel. 

\section{Results and Discussion}\label{sec:results_analysis}

\subsection{Main Performance Comparison}
To the best of our knowledge, this work is the first to utilize the PM25Vision dataset. For the TRAQID dataset, existing studies predominantly focus on centralized approaches, with limited exploration in decentralized scenarios. To enable a rigorous evaluation, we construct a comprehensive set of unimodal and multimodal baselines in addition to existing methods. The unimodal baselines include a Multilayer Perceptron ($\mathcal{M}$) for sensor data and lightweight CNN architectures such as MobileNetV3-Small, MobileNetV2, and ShuffleNetV2 x0.5 for image-based downstream task. The multimodal baselines are formed by combining these CNN backbones with the tabular MLP ($\mathcal{M}$), namely MobileNetV3-Small + $\mathcal{M}$, MobileNetV2 + $\mathcal{M}$, and ShuffleNet-V2x0.5 + $\mathcal{M}$, where the ‘+’ operator denotes simple extracted feature concatenation. The CNN backbones are kept frozen during training with a trainable classification head.

To ensure a fair and rigorous evaluation, we compare M$^2$FedAQI under both centralized and federated settings. In the centralized setting, we evaluate against existing methods in addition to the standard unimodal and multimodal baselines constructed in this work. As there are no prior federated learning benchmarks available for the considered datasets, we conduct comparisons in the federated setting using the same set of standardized baselines. This unified evaluation protocol enables a consistent and comprehensive assessment across both learning paradigms. Tables~\ref{tab:centralized_results} and \ref{tab:federated_results} present the results for the centralized and federated settings, respectively.

\begin{figure*}[!h]
\centering
\subfloat[PM25: Classification]{
    \includegraphics[width=0.235\textwidth]{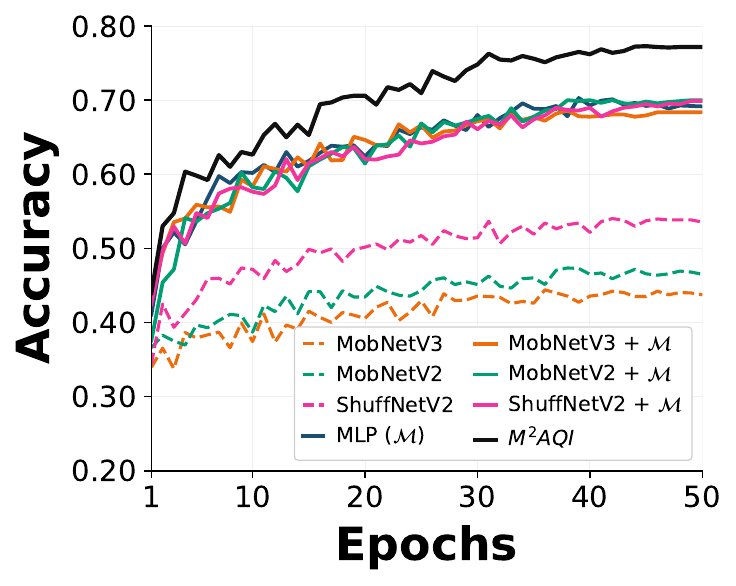}
}
\hfill
\subfloat[PM25: Regression]{
    \includegraphics[width=0.235\textwidth]{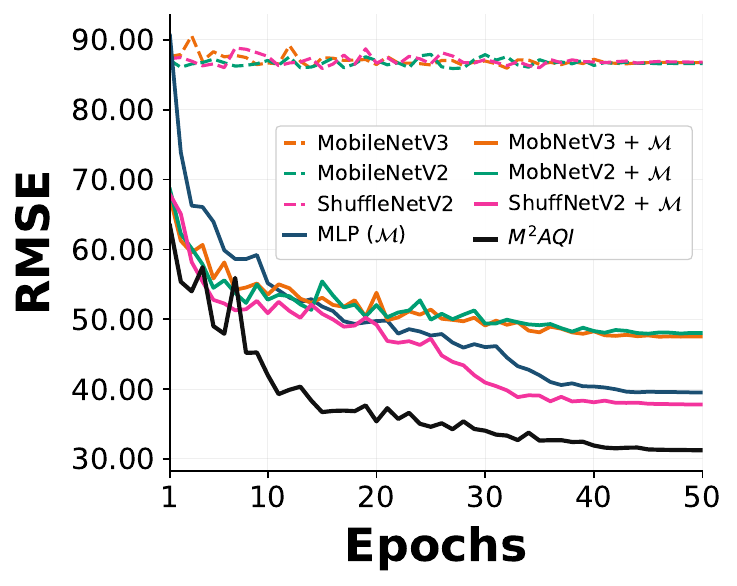}
}
\hfill
\subfloat[TRAQID: Classification]{
    \includegraphics[width=0.235\textwidth]{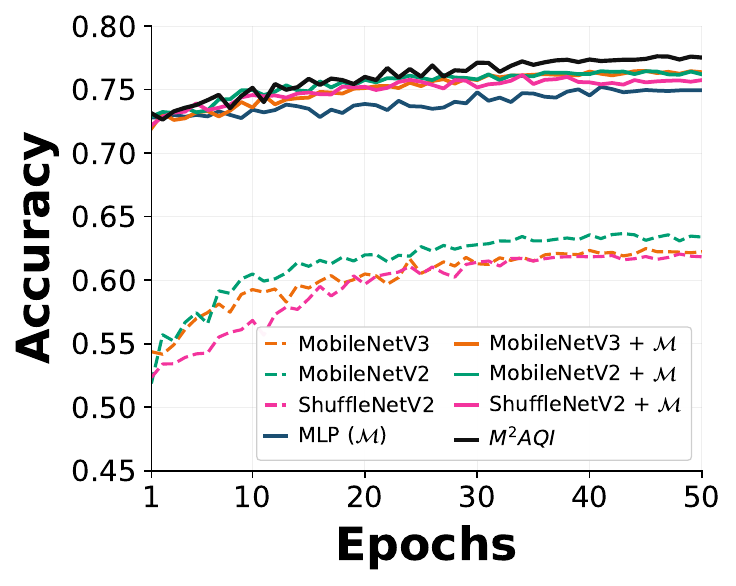}
}
\hfill
\subfloat[TRAQID: Regression]{
    \includegraphics[width=0.235\textwidth]{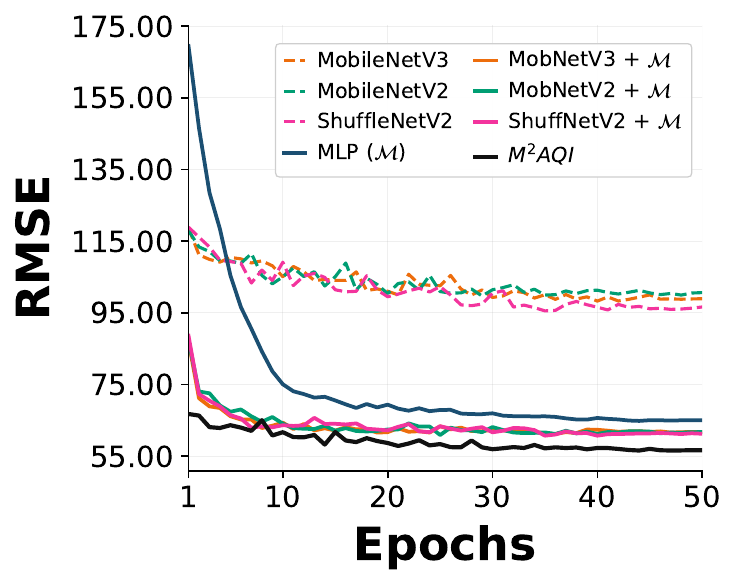}
}
\caption{Centralized Performance for PM25Vision \& TRAQID}
\label{fig:cent_line_plots}
\end{figure*}

\begin{figure*}[!h]
\centering
\subfloat[PM25: Classification]{
    \includegraphics[width=0.235\textwidth]{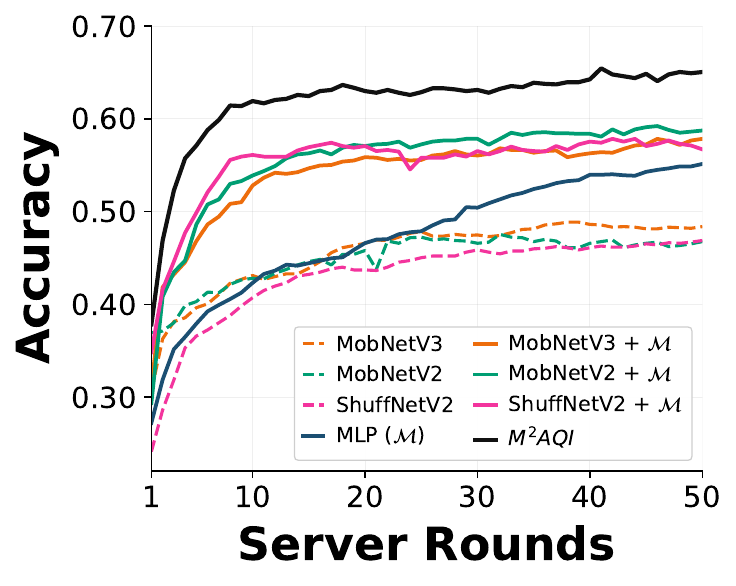}
}
\hfill
\subfloat[PM25: Regression]{
    \includegraphics[width=0.235\textwidth]{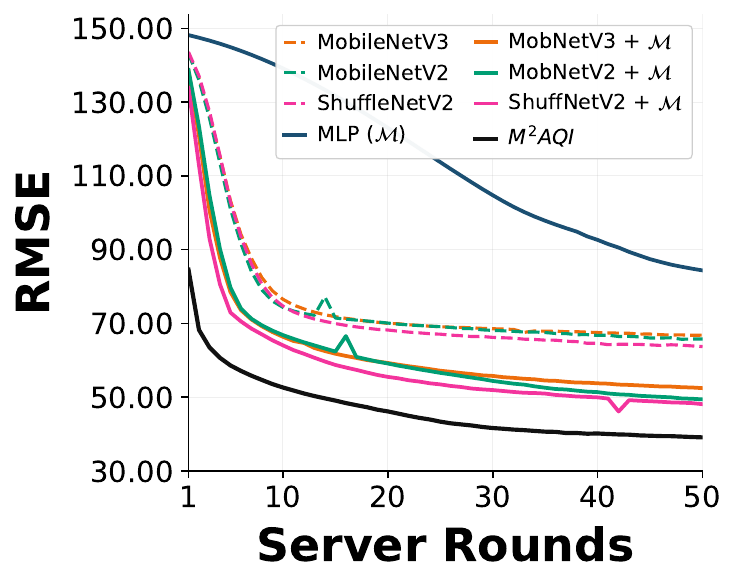}
}
\hfill
\subfloat[TRAQID: Classification]{
    \includegraphics[width=0.235\textwidth]{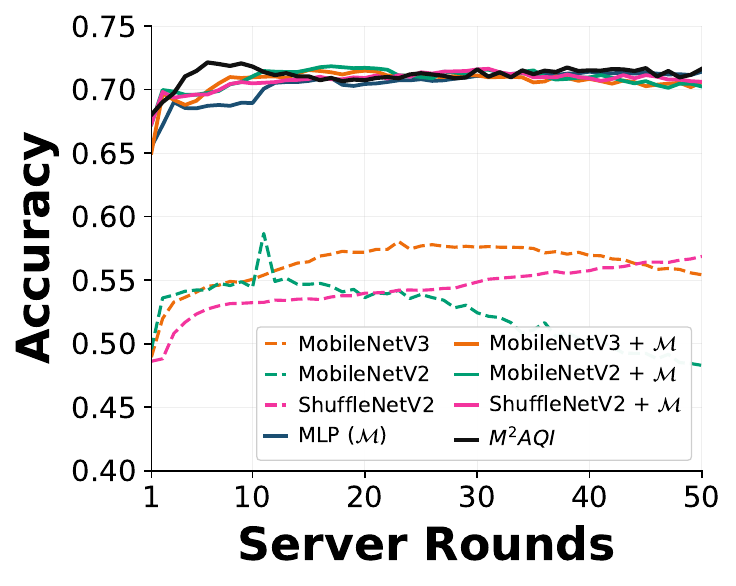}
}
\hfill
\subfloat[TRAQID: Regression]{
    \includegraphics[width=0.235\textwidth]{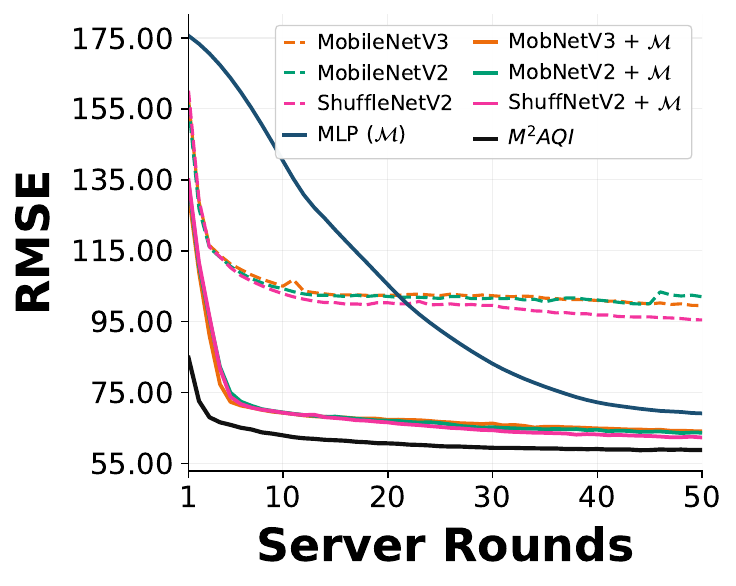}
}
\caption{Federated Performance for PM25Vision \& TRAQID}
\label{fig:fed_line_plots}
\end{figure*}

In addition, the training curves for the centralized and federated settings are shown in Fig.~\ref{fig:cent_line_plots} and Fig.~\ref{fig:fed_line_plots}, respectively. The plots demonstrate consistent convergence behavior, with steady performance improvement across training epochs in the centralized setting and communication rounds in the federated setting.

\begin{table*}[!h]
{\fontsize{7.5}{10}\selectfont
\caption{Centralized Performance Comparison ($\times$: Not Applicable \quad N/A: Not Available \quad $\mathcal{M}$: MLP)}
\label{tab:centralized_results}
\renewcommand{\arraystretch}{1.05}
\begin{tabular}{|p{3.0cm}|p{0.7cm}|p{0.68cm}p{0.85cm}p{0.69cm}|p{0.84cm}p{0.98cm}p{0.61cm}|p{0.68cm}p{0.85cm}p{0.60cm}|p{0.84cm}p{0.98cm}p{0.59cm}|}
\hline
\rowcolor{headergray}
\textbf{Method} & \textbf{Modal} 
& \multicolumn{6}{c|}{\textbf{PM25Vision}} 
& \multicolumn{6}{c|}{\textbf{TRAQID}} \\
\hline
 &  & \multicolumn{3}{c|}{Classification} & \multicolumn{3}{c|}{Regression}
 & \multicolumn{3}{c|}{Classification} & \multicolumn{3}{c|}{Regression} \\
\hline
 &  & Acc $\uparrow$ & AUC $\uparrow$ & F1 $\uparrow$ 
 & MAE $\downarrow$ & RMSE $\downarrow$ & $R^2$ $\uparrow$
 & Acc $\uparrow$ & AUC $\uparrow$ & F1 $\uparrow$
 & MAE $\downarrow$ & RMSE $\downarrow$ & $R^2$ $\uparrow$ \\
\hline

MLP ($\mathcal{M}$) & Tabular 
& \underline{0.693} & 0.906 & \underline{0.682} & 24.54 & 39.51 & 0.798
& 0.751 & 0.917 & 0.732 & 35.63 & 66.65 & \underline{0.645} \\

MobileNetV3-Small & Image 
& 0.425 & 0.766 & 0.411 & 66.72 & 87.89 & -5e-4 
& 0.619 & 0.796 & 0.557 & 62.06 & 101.16 & 0.182\\

MobileNetV2 & Image 
& 0.476 & 0.792 & 0.477 & 66.75 & 87.89 & -3e-4 
& 0.626 & 0.808 & 0.584 & 63.11 & 102.60 & 0.158 \\

ShuffleNet V2 x0.5 & Image 
& 0.529 & 0.828 & 0.522 & 66.69 & 87.89 & -6e-4 
& 0.610 & 0.792 & 0.547 & 61.06 & 98.42 & 0.225 \\

MobileNetV3-Small + $\mathcal{M}$ & Both 
& 0.679 & \underline{0.907} & 0.668 & 28.43 & 54.48 & 0.615 
& 0.756 & \underline{0.924} & 0.737 & 34.14 & 64.07 & 0.672  \\

MobileNetV2 + $\mathcal{M}$ & Both 
& 0.682 & 0.902 & 0.664 & 28.03 & 55.29 & 0.604 
& 0.754 & 0.920 & 0.737 & 35.04 & 62.89 & {0.684}  \\

ShuffleNet V2 x0.5 + $\mathcal{M}$ & Both 
& 0.679 & 0.900 & 0.654 & \underline{23.90} & \underline{38.49} & \underline{0.808}  
& \underline{0.760} & 0.922 & \underline{0.740} & 34.07 & 63.32 & 0.679 \\

Yang Han~\cite{han2025pm25vision} & Image 
& 0.440 & N/A & 0.380 & 36.60 & 54.60 & 0.550 
& $\times$ & $\times$ & $\times$ & $\times$ & $\times$ & $\times$ \\

Mondal et al. ~\cite{mondal2024uncovering} & Both 
& $\times$ & $\times$ & $\times$ & $\times$ & $\times$ & $\times$
& 0.640 & N/A  & 0.610 & 36.69 & 60.34 & N/A \\

Kalajdjieski et al. ~\cite{kalajdjieski2020air} & Both 
& $\times$ & $\times$ & $\times$ & $\times$ & $\times$ & $\times$
& 0.600 & N/A  & 0.560 & 41.07 & 68.28 & N/A \\

Nilesh et al. ~\cite{nilesh2022iot} & Both 
& $\times$ & $\times$ & $\times$ & $\times$ & $\times$ & $\times$
& 0.730 & N/A & 0.710 & 33.19 & \underline{54.24} & N/A \\

AQC-Net ~\cite{zhang2020deep}  & Both 
& $\times$ & $\times$ & $\times$ & $\times$ & $\times$ & $\times$
& 0.750 & N/A  & 0.740 & \underline{31.67} & \textbf{52.21} & N/A \\
  
\textbf{M$^2$FedAQI} & Both
& \textbf{0.769} & \textbf{0.939} & \textbf{0.765} & \textbf{17.83} & \textbf{30.62} & \textbf{0.879} 
& \textbf{0.765} & \textbf{0.935} & \textbf{0.753} & \textbf{31.66} & {60.53} & \textbf{0.707}  \\
\hline
\texttt{Improvement(\%)} & 
& \texttt{11.0} & \texttt{3.53}  & \texttt{12.2} 
& \texttt{25.4} & \texttt{20.4} & \texttt{8.79} 
& \texttt{0.66} & \texttt{1.19} & \texttt{1.76} 
& \texttt{0.02} & \texttt{-15.9} & \texttt{9.61} \\
\hline
\end{tabular}
}
\end{table*}

\begin{table*}[!h]
{\fontsize{7.5}{10}\selectfont
\caption{Federated Performance Comparison ($\mathcal{M}$: MLP)}
\label{tab:federated_results}
\renewcommand{\arraystretch}{1.05}
\begin{tabular}{|p{3.0cm}|p{0.7cm}|p{0.68cm}p{0.85cm}p{0.69cm}|p{0.84cm}p{0.98cm}p{0.59cm}|p{0.68cm}p{0.85cm}p{0.60cm}|p{0.84cm}p{0.98cm}p{0.59cm}|}
\hline
\rowcolor{headergray}
\textbf{Method} & \textbf{Modal} 
& \multicolumn{6}{c|}{\textbf{PM25Vision}} 
& \multicolumn{6}{c|}{\textbf{TRAQID}} \\
\hline
 &  & \multicolumn{3}{c|}{Classification} & \multicolumn{3}{c|}{Regression}
 & \multicolumn{3}{c|}{Classification} & \multicolumn{3}{c|}{Regression} \\
\hline
 &  & Acc $\uparrow$ & AUC $\uparrow$ & F1 $\uparrow$ 
 & MAE $\downarrow$ & RMSE $\downarrow$ & $R^2$ $\uparrow$
 & Acc $\uparrow$ & AUC $\uparrow$ & F1 $\uparrow$
 & MAE $\downarrow$ & RMSE $\downarrow$ & $R^2$ $\uparrow$ \\
\hline

MLP ($\mathcal{M}$) & Tabular  & 0.551 & 0.879 & 0.593 & 57.47 & 84.37 & 0.014
& 0.714 & \underline{0.914} & 0.715 & 38.24 & 69.09 & 0.516 \\

MobileNetV3-Small & Image 
& 0.489 & 0.836 & 0.542 & 45.01 & 66.77 & 0.340
& 0.580 & 0.765 & 0.573 & 59.22 & 99.52 & 0.044 \\

MobileNetV2 & Image  & 
0.475 & 0.816 & 0.523 & 44.54 & 65.69 & 0.370
& 0.587 & 0.738 & 0.567 & 59.65 & 99.96 & 0.036 \\

ShuffleNet V2 x0.5 & Image 
& 0.469 & 0.819 & 0.515 & 43.55 & 63.72 & 0.394
& 0.569 & 0.764 & 0.539 & 57.26 & 95.46 & 0.109 \\

MobileNetV3-Small + $\mathcal{M}$ & Both 
& 0.578 & \underline{0.895} & 0.632 & 33.13 & 52.45 & 0.565
& 0.715 & 0.913 & 0.727 & 35.37 & 64.06 & 0.600 \\

MobileNetV2 + $\mathcal{M}$ & Both 
& \underline{0.592} & 0.893 & \underline{0.642} & 34.07 & 49.39 & 0.620
& \underline{0.718} & 0.908 & 0.730 & 35.56 & 63.56 & 0.602 \\

ShuffleNet V2 x0.5 + $\mathcal{M}$ & Both 
& 0.578 & 0.886 & 0.623 & \underline{31.56} & \underline{46.11} & \underline{0.634}
& 0.716 & 0.914 & \underline{0.733} & \underline{35.06} & \underline{62.29} & \underline{0.619} \\

\textbf{M$^2$FedAQI} & Both
& \textbf{0.649} & \textbf{0.919}  & \textbf{0.691} 
& \textbf{26.38} & \textbf{39.09} & \textbf{0.748} 
& \textbf{0.721} & \textbf{0.921} & \textbf{0.737} 
& \textbf{33.27} & \textbf{58.71} & \textbf{0.654} \\
\hline
\texttt{Improvement(\%)} & 
& \texttt{9.63} & \texttt{2.68}  & \texttt{7.63} 
& \texttt{16.4} & \texttt{15.2} & \texttt{18.0} 
& \texttt{0.42} & \texttt{0.77} & \texttt{0.55} 
& \texttt{5.11} & \texttt{5.75} & \texttt{5.65} \\
\hline
\end{tabular}
}
\end{table*}

\subsection{Edge Profiling of $M^2FedAQI$}
\begin{figure}[H]
    \centering
    \subfloat[]{
        \includegraphics[width=0.23\textwidth]{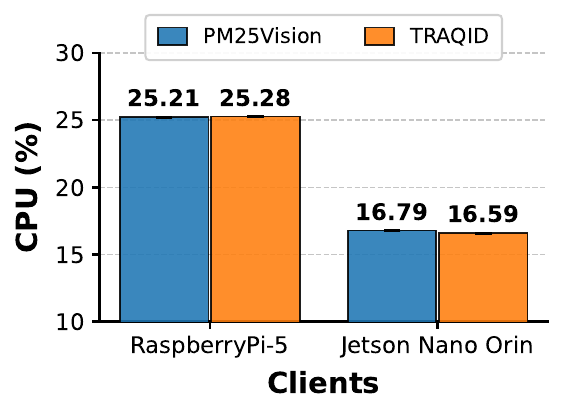}
    }
    \hfill
    \subfloat[]{
        \includegraphics[width=0.23\textwidth]{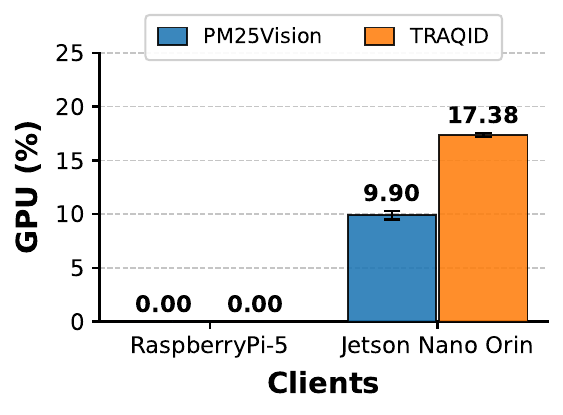}
    }
    \hfill
    \subfloat[]{
        \includegraphics[width=0.23\textwidth]{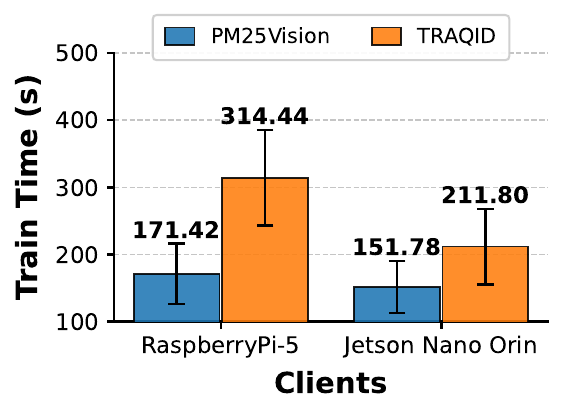}
    }
    \subfloat[]{
        \includegraphics[width=0.23\textwidth]{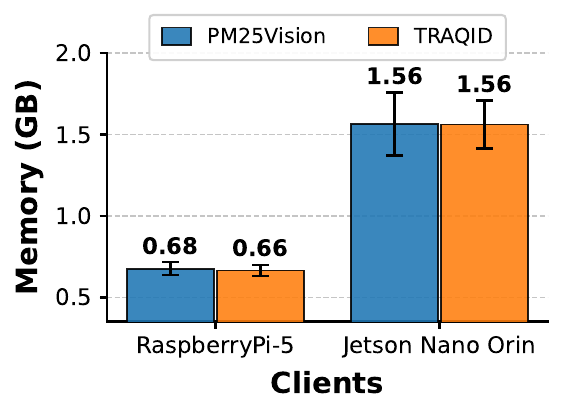}
    }
    \hfill
    \subfloat[]{
        \includegraphics[width=0.23\textwidth]{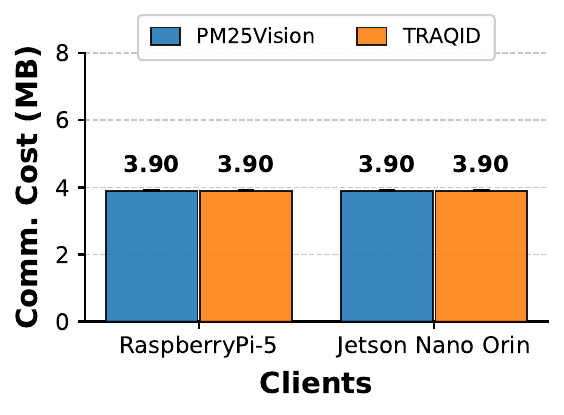}
    }
    \caption{Federated Edge Profiling of M$^2$FedAQI}
    \label{fig:edge_profile}
\end{figure}

Fig.~\ref{fig:edge_profile} presents the federated edge profiling of M$^2$FedAQI across Raspberry Pi~5 and NVIDIA Jetson Orin Nano devices for both PM25Vision and TRAQID datasets. The profiling evaluates CPU utilization (\%), GPU utilization (\%), local training time per communication round (s), memory footprint (GB), and communication cost during model upload (MB).

The results indicate that CPU utilization remains moderate across both devices, demonstrating that the proposed model does not impose excessive computational load. As expected, GPU utilization is observed only on the Jetson platform, while the Raspberry Pi~5 operates entirely on CPU. The local training time per round remains within practical limits on both devices, reflecting the efficiency of the lightweight architecture. Furthermore, the memory footprint is consistently low, confirming that M$^2$FedAQI is well-suited for deployment on resource-constrained edge hardware.

Overall, these findings highlight the practical feasibility of deploying M$^2$FedAQI in real-world heterogeneous edge environments. Unlike most existing works that focus solely on predictive performance, this study provides a comprehensive system-level evaluation, demonstrating that the proposed approach achieves a favorable balance between accuracy and computational efficiency.

\subsection{Ablation Study}
\begin{figure}[H]
    \centering
    \subfloat[PM25Vision: Classification]{
        \includegraphics[width=0.2305\textwidth]{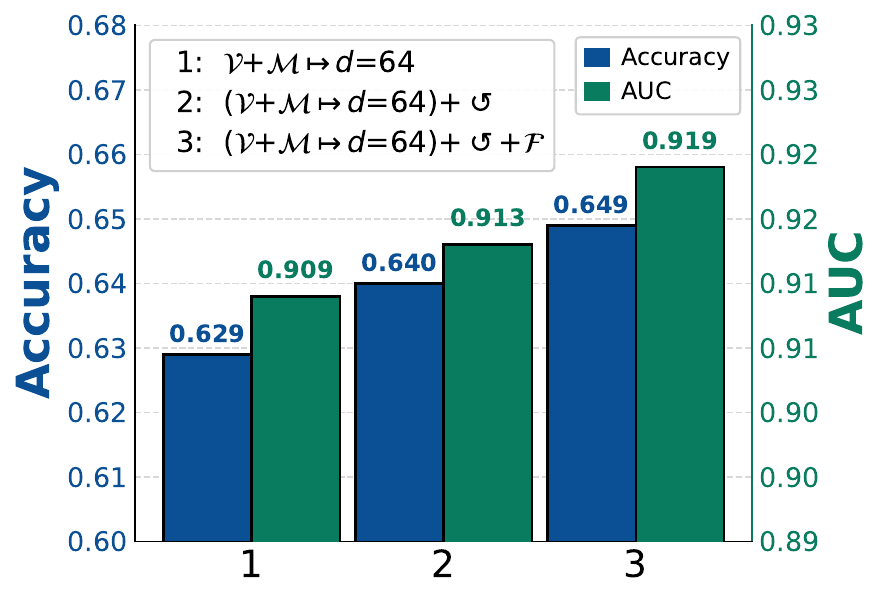}
    }
    \hfill
    \subfloat[PM25Vision: Regression]{
        \includegraphics[width=0.2305\textwidth]{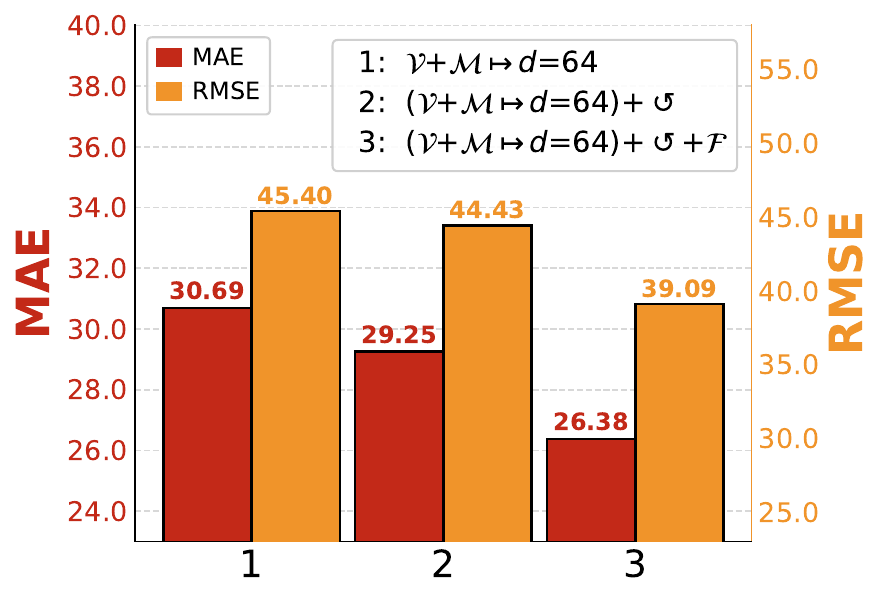}
    }
    \hfill
    \subfloat[TRQAID: Classification]{
        \includegraphics[width=0.2305\textwidth]{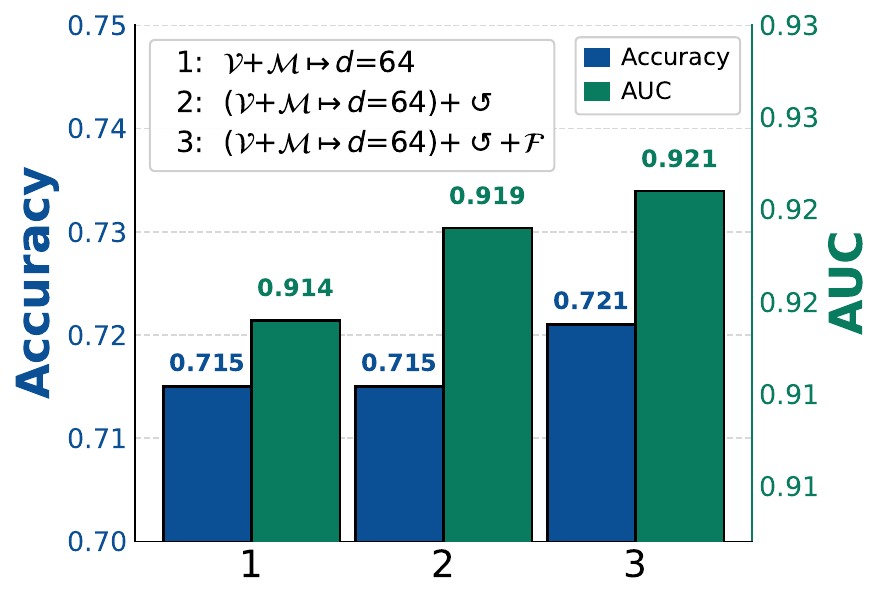}
    }
    \hfill
    \subfloat[TRQAID: Regression]{
        \includegraphics[width=0.2305\textwidth]{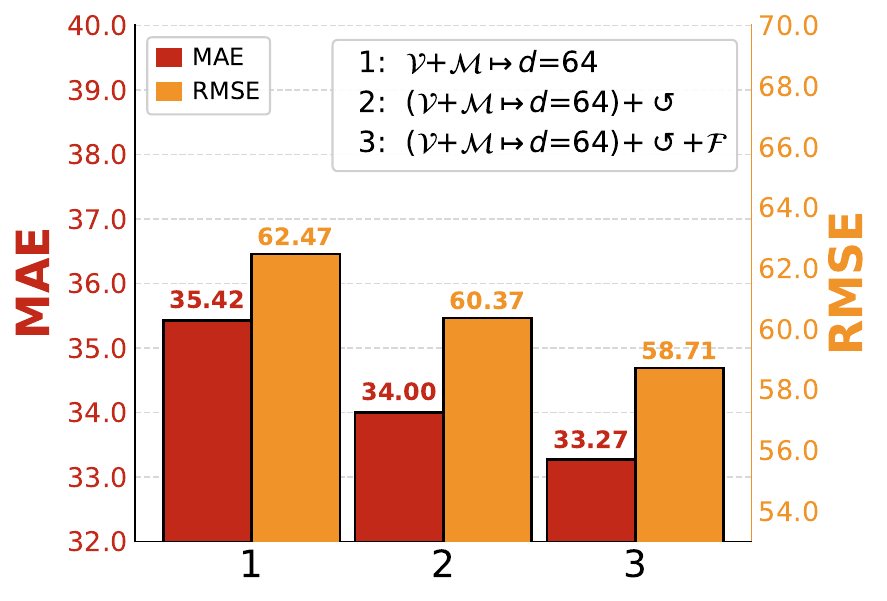}
    }
    \caption{Ablation Study of M$^2$FedAQI }
    \label{fig:fed_ablation}
\end{figure}

To evaluate the contribution of key architectural components, we conduct an ablation study by incrementally incorporating skip connections and the proposed fusion mechanism, as shown in Fig.~\ref{fig:fed_ablation}. 

\subsubsection{Effect of Skip Connections ($\circlearrowleft$):}
Introducing residual skip connections in the tabular branch improves feature preservation and stabilizes training. This results in consistent gains across both classification and regression tasks, indicating better utilization of structured environmental features. 

\subsubsection{Effect of Fusion Module ($\mathcal{F}$):}
The proposed feature modulation–based fusion mechanism enables effective interaction between visual and tabular modalities. Compared to simpler integration strategies, this leads to notable improvements in predictive performance, highlighting the importance of cross-modal feature alignment. 

\subsubsection{Combined Architecture ($\circlearrowleft + \mathcal{F}$):}
Combining skip connections with the fusion module achieves the best overall performance across all datasets and tasks. The results demonstrate that both components are complementary, jointly enhancing representation quality and multimodal learning. 

\subsubsection{Overall Observation}
The ablation study confirms that each component contributes to incremental performance gains, with the fusion mechanism providing the most significant improvement.

\section{Conclusion and Future Directions}
\label{sec:conclusion}
This work presents a lightweight multimodal approach for decentralized air quality prediction by effectively integrating visual and tabular data through a feature modulation--based fusion mechanism. The proposed model is designed for resource-constrained environments, enabling efficient cross-modal learning while maintaining a low computational footprint. Experimental results on PM25Vision and TRAQID demonstrate that M$^2$FedAQI consistently outperforms unimodal and baseline multimodal methods across both regression and classification tasks under centralized and federated settings. Beyond predictive performance, the proposed approach is validated through deployment on heterogeneous edge devices, with detailed analysis of latency, memory usage, computational efficiency, and communication cost. Secure client participation is ensured through TLS-based authentication integrated within the standard federated learning pipeline without modifying the underlying FL protocol. 

Overall, the results demonstrate the effectiveness, scalability, and practicality of combining lightweight multimodal learning with federated learning for privacy-preserving air quality prediction in real-world IoT environments. Future work will focus on enabling personalized federated learning to better handle client-specific data heterogeneity, improving communication efficiency for large-scale real-world deployments, and incorporating richer multimodal data sources such as satellite imagery, weather conditions, and traffic information for more comprehensive and robust AQI prediction.

\footnotesize
\section*{Acknowledgments}\label{sec:acknowledgements}
This work was partially supported by the Indian Institute of Technology, (ISM) Dhanbad via grant no. MISC0203

\normalsize
\bibliography{references}

\clearpage


\end{document}